\documentclass[10pt,twocolumn,letterpaper]{article}

\usepackage{cvpr} % uncomment this for the final submission

\usepackage{times}
\usepackage{graphicx}
\usepackage[norule,symbol,perpage]{footmisc}

\usepackage{epsfig}
\usepackage{multirow,tabularx}
\usepackage{amsmath,amssymb}
\usepackage{ifthen}
\usepackage{color}
\usepackage{xcolor}
\usepackage{booktabs}
\usepackage{afterpage}
\usepackage{subcaption}
\DeclareGraphicsExtensions{.pdf,.jpg,.png}
\graphicspath{{./graphics/}{./figures/}{./results/}}

\newcolumntype{C}{X<{\centering}}

\definecolor{lightgreen}{rgb}{0.67, 0.88, 0.69}
\definecolor{darkgreen}{rgb}{0,0.55,0}
\definecolor{linkcolor}{rgb}{0,0,.65}

\newcommand\Caption[3][]{\caption[#2]{\label{#1}\textsc{#2}. \small#3}}

\newcommand{\figref}[1]{Fig.~\ref{#1}\xspace}

\newcommand{\tabref}[1]{Tab.~\ref{#1}\xspace}

\newcommand{\ie}{\textit{i.\,e.}\xspace}
\newcommand{\eg}{\textit{e.\,g.}\xspace}

\newcommand{\etal}[1]{\textit{et al.}~\cite{#1}\xspace}

\usepackage[pagebackref=true,breaklinks=true,letterpaper=true,colorlinks,bookmarks=false]{hyperref}

\begin{document}

%%%%%%%%% TITLE

\title{Bias Mitigation in Face Recognition via Demographic-based Supervised Contrastive Learning}

%%%%%%%%% AUTHORS - PLEASE UPDATE
\author{
Yu Linghu, Salman Mohammad, Xinyi Zhang, Manuel G\"unther\\
Department of Informatics\\
University of Zurich\\
\texttt{\small \{yu.linghu,salman.mohammad,xinyi.zhang,manuel.guenther\}@uzh.ch}
}

\maketitle
\thispagestyle{empty}

%\switchlinenumbers*

\begin{abstract}
Face recognition systems have been shown to be biased toward certain demographic groups by exhibiting different error rates across gender, age, or ethnicity.
Though the imbalance of the training data with respect to these demographics is one cause of this bias, training on artificially balanced groups does not completely mitigate the problem.
For deployment, face recognition typically works at operating points allowing very low false match rates and, hence, on the tail of the non-match score distribution.
While class balancing can improve the means of these distributions, the aim of our approach is to improve fairness by addressing the behavior in the tail.
Particularly, we propose the Demographic-based Supervised Contrastive loss (DeSCon) for face recognition, which relies on a well-designed composition of training batches and demographic-aware pair selection.
Our experimental evaluation on both demographically-labeled datasets and standard verification benchmarks shows that DeSCon can improve fairness beyond balancing training datasets while maintaining competitive verification performance.
Source code is available upon request.
\end{abstract}

\section{Introduction}
\label{sec:introduction}

Face recognition has become one of the most widely deployed biometric technologies, underpinning automatic verification and identification processes across a broad range of applications.
Particularly, automatic border control systems \cite{hidayat2024abc} reach performances beyond human capabilities \cite{otoole2007surpass}.
Also, face recognition is utilized for unlocking mobile devices \cite{kunda2021android}, or in surveillance systems \cite{kasim2024challenge}.

Especially approaches based on deep learning have provided a huge performance gain over traditional face recognition systems \cite{pereira20228years}.
In general, a deep learning-based face recognition system exploits the network as a feature extractor.
For a given facial image, the face is detected, aligned, and input to the deep network, which extracts a face embedding, also called a face template, from the image.
% Such a template is stored on a passport or mobile device during regular verification tasks.
Such a template is stored on a mobile device during enrollment for later verification.
For travel documents, a facial image is stored on the passport chip. 
In order to verify one's identity, another live photograph is taken, an embedding is extracted, and the similarity to the stored template is computed.
If this similarity score exceeds a pre-defined threshold, the verification is successful; otherwise, the probe face is rejected.

However, such networks are trained on large facial image datasets crawled from the internet, which are typically imbalanced in terms of demographics that they capture \cite{robbins2023cast}.
It has been observed that this imbalance is translated into different performances of these systems for different demographic groups \cite{cavazos2020accuracy}, \eg, such bias can lead to different treatment of people with different skin colors \cite{hill2020wrongfully}.

\begin{figure}[t]
  \includegraphics[width=\linewidth]{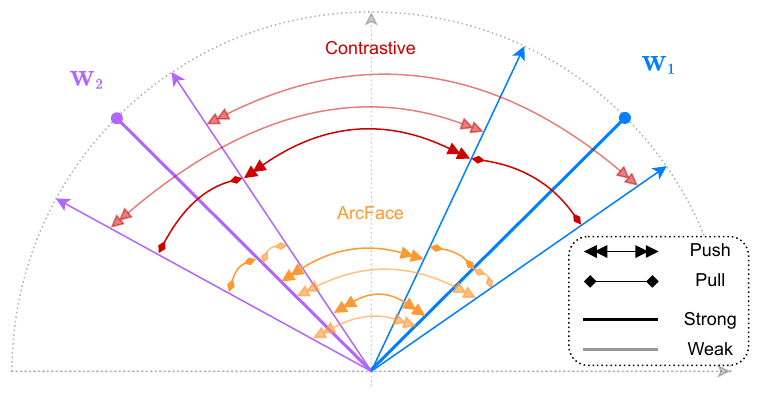}
  \Caption[fig:teaser]{Centroid-Based and Contrastive Learning}{
    When training with centroid-based margin losses such as ArcFace, feature vectors are drawn to their own class center $\mathbf W_y$ and pushed away from other class centers.
    However, for deployment, only embeddings are compared.
    Supervised contrastive loss works on the basis of embeddings directly, pulling the mated same-identity embeddings (same color) together and pushing the non-mated different-identity embeddings (different colors) away.
    Due to the definition of SoftMax in the contrastive loss \eqref{eq:contrastive}, the force for nearby different-identity features is much stronger; thus, contrastive learning works directly on the tail of the non-mated score distributions, thereby making systems more fair in operational settings without impeding overall verification performance.
    In DeSCon, demographic labels guide the selection of non-mated pairs, restricting or prioritizing sampling within the same demographic group to specifically target within-group separation.}
\end{figure}

To mitigate such bias, two main approaches have been developed.
The first and simpler approach tries to handle bias through post-hoc analysis of verification scores, by defining separate thresholds per demographic group  \cite{cook2019demographic,poh2010group}, or using score normalization techniques to align score distributions across demographic groups \cite{terhorst2020post,linghu2024score}.
Unfortunately, such techniques often require demographic information during deployment, and do not increase the overall performance of the system.

More research is dedicated to the development of algorithms for training deep networks to extract more fair features \cite{gong2020jointly,liang2019additive,morales2020sensitivenets}, or adapting pre-trained networks for providing better-aligned score distributions \cite{kotwal2024mitigating}.
Many of the former methods remove sensitive information about current demographics from the embeddings -- when the embedding does not include information about gender, it can not be biased against it.
However, since such demographic information is very useful in the recognition process, such systems often drop overall verification performance \cite{gong2020jointly,morales2020sensitivenets}.
Another option is to balance the training data with respect to the demographics.
Unfortunately, as has been reported in the literature \cite{albiero2020balance,kolla2023impact} and is also verified in our experiments, only balancing the training is not sufficient to arrive at the most fair system, while it can additionally result in reduced overall verification performance.

In this work, we propose a new Demographic-based Supervised Contrastive loss (DeSCon) for fairness-aware face recognition.
% We integrate supervised contrastive learning \cite{khosla2020supervised}, which works by directly comparing embeddings of two training samples within the same demographic group, into margin-based softmax frameworks.
We integrate supervised contrastive learning \cite{khosla2020supervised}, which works by directly comparing embeddings of training samples through pairwise supervision, into margin-based softmax frameworks.
As visualized in \figref{fig:teaser}, our core idea is to complement the class-level pull of class center-based losses such as ArcFace \cite{deng2022arcface} with a structured push against non-mated pairs, thereby reducing demographic disparities while preserving discriminative power.
We design three selection strategies to explore this principle:
(1) \emph{DeSCon-WG} restricts non-mated pair sampling within demographic groups to enhance local balance, while
(2) \emph{DeSCon-Hard} further focuses on the hardest non-mated pairs to maximize fairness regularization and moves further into the tail of the non-match score distribution.
To show the impact of our demographic-based pair selection strategy, we also compare to (3) \emph{DeSCon-All} that applies contrastive loss to non-mated pairs within and across demographic groups.

We validate our methods through extensive experiments on multiple benchmarks and fairness metrics.
The results demonstrate that DeSCon improves True Match Rates (TMR) while reducing demographic disparities in False Match Rates (FMR) and False Non-Match Rates (FNMR) across a range of evaluation settings.
Our analysis further reveals that different DeSCon variants are preferable under different data regimes:
Though DeSCon-All is effective when training data are balanced, DeSCon-Hard exhibits greater robustness when training or evaluation data are demographically imbalanced, delivering the most consistent fairness improvements over the baseline.
% Overall, DeSCon outperforms state-of-the-art preprocessing and in-processing baselines across a wide range of evaluation settings.
Overall, DeSCon achieves a competitive balance between verification performance and fairness, with DeSCon-Hard providing the most robust fairness improvements across the evaluated settings.

\noindent As our contributions in this paper, we:
\begin{itemize}
    \item introduce DeSCon, a fairness-aware face recognition loss that augments centroid-based losses with supervised contrastive learning to jointly enhance accuracy and fairness,
    \item explore three pair sampling strategies (All, WG, Hard) to balance between global separation, within-group consistency, and hard non-mated regularization,
    \item conduct comprehensive evaluations across multiple backbones and benchmarks, showing that DeSCon-Hard consistently improves fairness over the baseline while remaining competitive with existing bias-mitigation methods in terms of the fairness-performance trade-off, and
    \item evaluate state-of-the-art methods using both conventional and novel evaluation metrics, highlighting the need to comply with metrics defined in international standards.
\end{itemize}

\section{Related Work}
\label{sec:related}

\subsection{Training Face Recognition Networks}
In recent years, deep learning has dominated and revolutionized many fields of research, including Face Recognition (FR).
In general, there exist two main directions: developing better network topologies and implementing better-suited loss functions.
Most modern network architectures include variations and improvements of residual network architectures \cite{he2016deep,hu2018squeeze,duta2021iresnet}, and vision transformers \cite{sun2022part}.
The latest developed loss functions, \ie, ArcFace \cite{deng2022arcface}, MagFace \cite{meng2021magface}, and AdaFace \cite{kim2022adaface} improve the discriminability of deep features in angular space by comparing face embeddings with a class centroid.
Many of these networks are trained on huge amounts of data \cite{zhu2021webface260m}, mainly capturing celebrities, and exhibiting a large imbalance in the distribution of demographic groups.

\subsection{Fairness Issues in Face Recognition}
A few years ago, the performance of FR systems across demographic groups gained news coverage \cite{hill2020wrongfully}, where it was shown that a commercial-off-the-shelf algorithm performed much poorer on people of color than on the white population.
This triggered research in this domain, and it was shown that several demographic factors, such as ethnicity \cite{vangara2019characterizing,krishnapriya2020skintone}, gender \cite{albiero2020gender}, age \cite{michalski2018impact}, eyewear \cite{cook2019demographic}, or even the amount of facial hair \cite{wu2024facial-hair}, influence face recognition.
In these works, several datasets have been exploited to perform the fairness evaluation, including MORPH \cite{pereira2021fairness}, Racial Faces in the Wild (RFW) \cite{wang2019racial}, BUPT \cite{wang2020mitigating}, DiveFace \cite{morales2020sensitivenets}, and Balanced Faces in the Wild (BFW) \cite{robinson2020face,robinson2023balancing}.
Datasets like UTKFace \cite{zhang2017age} and FairFace \cite{karkkainen2021fairface} also serve the same purpose without ground-truth labels for identities.

An important aspect relates to measuring fairness across demographic groups.
While earlier work simply split the datasets into different demographics and reported separate ROC curves per demographic \cite{huang2019deep,kortylewski2019analyzing,wang2020mitigating,wang2019racial,wang2019deep,krishnapriya2020skintone}, it has been pointed out that such evaluation does not align with real-world deployment that requires a single global operational threshold.
Subsequently, several metrics have been developed, but the National Institute of Standards and Technology (NIST) concludes that none of the existing metrics can cover all fairness aspects \cite{grother2022nist}.
Lately, ISO/IEC 19795-10 \cite{iso19795-10} specifies how to measure and report fairness performance.
Yet, the majority of works still report face verification performance using per-group accuracy -- exploiting demographic-specific thresholds -- rather than comparing TMR at a single operational threshold.

A simple yet effective way of improving the fairness of existing systems is a post-hoc score normalization \cite{wallace2012cross,mandasari2014calibration}.
For example, Linghu \etal{linghu2024score} investigate score normalization methods and extend these to demographics-based normalization techniques.
However, such systems need to compute additional similarities to cohort samples during enrollment or probing, and they need to know or estimate the demographics of a given gallery or probe sample.
A few more introspective approaches try to postprocess the original embeddings extracted by the deep network such that score distributions across demographics are more aligned \cite{kotwal2024mitigating}.
Other approaches include metric learning to improve distributions of scores over demographics \cite{terhoerst2020comparison,terhorst2020post}. 

\subsection{More Fair Network Training}

As it is proven that many facial attributes are encoded in face embeddings \cite{rudd2018hiding,terhorst2020beyond}, research focuses on learning less-biased face representations by disentangling the sensitive information from face embeddings \cite{gong2020jointly,liang2019additive,morales2020sensitivenets}, leveraging demographic labels available in the datasets.
Such works include cluster-based large-margin local embedding loss \cite{huang2019deep}, or reinforcement learning-based ethnicity-balanced networks \cite{wang2020mitigating}.
Yang \etal{yang2021ramface} propose to adjust the optimal margins for different ethnic groups.
Gong \etal{gong2021mitigating} develop a loss function that minimizes average intra-class distances between demographic groups.
Serna \etal{serna2022sensitive} extend triplet loss with a sensitive triplet generator to reduce discrimination, which works as an add-on to the pre-trained features.
Iurada \etal{iurada2024fairness} investigate debiasing through cross-domain learning.
Since demographic labels might be unreliable, Jung \etal{jung2022learning} propose Confidence-based Group Label assignment (CGL) of pseudo labels.
Other approaches include a false positive rate penalty loss \cite{xu2021consistent}, proxy features \cite{ardeshir2022estimating}, or minimizing overall variance of embeddings \cite{nagpal2022detox}.
Without relying on demographic labels, Wang \etal{wang2023mixfairface} reduce identity bias through feature mixing, while Ohki \etal{ohki2024labellessface} mitigate class favoritism via adaptive margins.

\subsection{Contrastive Learning}

Contrastive learning \cite{chen2020contrastive} is a self-supervised technique that uses a sample as an anchor, an augmented sample as a positive, and all other samples in the batch as negatives.
It has been used in self-supervised facial representation learning to separate pose-related from pose-unrelated factors \cite{liu2023pose}. 
In contrast to self-supervision, which assumes all samples in the batch to belong to different classes, Supervised Contrastive learning (SupCon) requires same-class samples within a batch \cite{khosla2020supervised} to pull same-class instances together and push different-class instances apart.
SupCon has been applied to several tasks on faces and beyond, \eg, for facial attribute classification \cite{ramachandran2024self,park2022fair}, or open-set face recognition \cite{ahn2024uncertainty,song2024coreface}, or demographics-aware medical image classification \cite{lin2024improving,du2022fairdisco}.
Additionally, various sampling strategies for obtaining different-class samples have been explored, including hard negative mining \cite{schroff2015facenet} and multi-similarity sampling \cite{wang2019multi}.
To the best of our knowledge, demographic-based sampling within SupCon for fairness-aware face recognition remains unexplored, and this work aims to fill this gap.

\section{Approach}
\label{sec:approach}

Our goal is to improve the consistency, separability, and fairness of face recognition embeddings by combining identity classification with contrastive supervision.
As illustrated in \figref{fig:teaser}, we leverage two complementary losses: a centroid-based loss such as ArcFace \cite{deng2022arcface} and SupCon \cite{khosla2020supervised}, to guide the embedding space jointly.
ArcFace enforces angular margins between identity prototypes, pulling each embedding toward its corresponding prototype.
SupCon, in parallel, encourages tight clustering of mated pairs while pushing apart embeddings from different identities, regardless of class boundaries.
The combination of classification-driven angular separation and pairwise contrastive forces leads to more robust and fair representations.

To support SupCon learning, we develop pair selection strategies that ensure the presence of mated pairs for each demographic group in each batch.
These carefully-sampled anchor-mated pairs enable stable contrastive optimization, allowing controlled exploration of non-mated pairs across identity and demographic boundaries.

\subsection{Identity Classification with ArcFace}
We follow the standard face recognition pipeline by training the model end-to-end using a classification objective over identities.
Given an input image, the network extracts an embedding $x \in \mathbb{R}^{M}$, which is passed to a classification head parameterized by a weight matrix $\mathbf{W} \in \mathbb{R}^{C \times M}$, with $C$ training identities.
To improve intra-class compactness and inter-class separability, we adopt the ArcFace loss \cite{deng2022arcface}, which introduces an angular margin in the hypersphere space.
This loss encourages compact and well-separated identity clusters on the unit hypersphere and serves as the foundation of our representation learning framework.
ArcFace applies an additive margin $m$ to the angle of target identity $y$:
\begin{equation}
    \label{eq:arcface}
    \mathcal{L}_{\mathrm{arc}}(x) = -\log \frac{e^{s \cdot \cos(\theta_y + m)}}{e^{s \cdot \cos(\theta_y + m)} + \sum\limits_{c \neq y} e^{s \cdot \cos(\theta_c)}}\,,
\end{equation}
where $\theta_c$ denotes the angle between the embedding $x$ and the class center $\mathbf{W}_c$.
The scale parameter $s$ controls the sharpness of the decision boundary, while the angular margin $m$ enforces stronger separation between identities \cite{zhang2019adacos}.

\subsection{Supervised Contrastive Learning}

To further improve representation consistency and separation between identities, we incorporate a Supervised Contrastive (SupCon) loss \cite{khosla2020supervised}.
SupCon learning leverages identity labels to define mated and non-mated pairs within a batch.
Given a batch of embeddings $\mathcal B = \{x_b\}_{b=1}^B$, associated with identity labels $y_b \in \{1, \dots, C\}$, the supervised contrastive loss for a single anchor embedding $x_a$ is \cite{khosla2020supervised}:
\begin{equation}
    \label{eq:contrastive}
    \hspace*{-.7em}\mathcal{L}_{\mathrm{sup}}(x_a) = - \frac{1}{|P(a)|} \sum_{x_p \in P(a)} \hspace*{-.6em} \log \frac{e^{s\cdot\cos(x_a,x_p)}}{\sum\limits_{x \in P(a) \cup N(a)} \hspace*{-1.3em} e^{s\cdot\cos(x_a,x)}}\hspace*{-.3em}
\end{equation}
where $P(a)\subset\mathcal B$ is the set of mated samples (same identity as $x_a$), and $N(a)\subset\mathcal B$ is the set of other samples from the batch, which we define below.
The scale parameter $s$, similarly to ArcFace, controls the softness of the distribution.
To ensure the availability of mated pairs for contrastive learning, we explicitly construct the training batch to contain mated pairs for each demographic group, sampled to cover the range of demographics uniformly.
We explore multiple strategies for selecting non-mated pairs $N(a)$ for computing the denominator in Eq.~\eqref{eq:contrastive}:
\begin{itemize}
    \item \textbf{All-vs-All (DeSCon-All):} all non-mated samples in the batch, regardless of demographics, 
    \item \textbf{Within-Group (DeSCon-WG):} non-mated samples belonging to the same demographic group as the anchor $x_a$, and 
    \item \textbf{Hardest Within-Group (DeSCon-Hard):} the top $K$ non-mated same-demographic samples that are most similar to $x_a$. 
\end{itemize}
These sampling strategies are designed to explore how different sources of non-mated pairs affect representation learning.
While DeSCon-All reflects the standard approach in contrastive learning, within-group and the hardest within-group selection specifically emphasize finer intra-demographic distinctions -- in our experiments, we use $K=10$.
By varying the non-mated pair selection strategy, we empirically assess whether more localized or demographically constrained contrastive supervision can improve fairness or robustness.

\subsection{Training Objective}

We train a model with a joint objective that combines ArcFace \eqref{eq:arcface} with SupCon \eqref{eq:contrastive}, both operating on the same normalized embedding space, allowing them to complement each other during training.
The demographics-weighted loss is:
\begin{equation}
    \mathcal{L}_{\text{total}} = \frac1B \sum\limits_{b=1}^B w_d \bigl[\mathcal{L}_{\mathrm{arc}}(x_b) + \lambda \cdot \mathcal{L}_{\mathrm{sup}}(x_b)\bigr]\,,
    \label{eq:joint}
\end{equation}
where $\lambda$ controls the relative contribution of the two objectives.
% In our experiments, we find that setting $\lambda = 1$ provides stable training and strong performance, although other values are explored in the ablation study.
In our experiments, we find that setting $\lambda = 1$ provides stable training and strong performance across all evaluated settings.
The two loss components are optimized jointly using standard stochastic gradient descent, without the need for staged training \cite{kotwal2024demographic}.

When the training dataset exhibits demographic imbalance, we incorporate group-aware loss weighting $w_d$  to compensate for unequal sample distributions \cite{palechor2023protocols}.
Specifically, each training sample is weighted according to its demographic group $d$ using:
\begin{equation}
    w_d = \frac N{D N_d}
\end{equation}
where $N$ is the total number of training samples, $D$ is the number of demographic groups, and $N_d$ denotes the number of samples belonging to group $d$.
When the training dataset is balanced across demographic groups, all weights reduce to $w_d = 1$, and the objective remains unchanged.

To maintain consistency with ArcFace \eqref{eq:arcface}, we use the same scale $s$ in the contrastive loss \eqref{eq:contrastive}.
The angular margin $m$ is applied only in ArcFace, as it is specific to classification-based separation and incompatible with the symmetric formulation of contrastive learning.

\section{Experiments}
\label{sec:experiments}

\subsection{Evaluation Metrics}
\label{sec:metrics}
Following common practice in face recognition, we compute a score threshold $\tau$ to arrive at a fixed False Match Rate (FMR), and we report the True Match Rate (TMR) at $\tau$ as our main metric for overall verification performance.

For fairness evaluation, we adopt two complementary evaluation setups to assess model performance and fairness.
As our primary evaluation metric, we follow the ISO/IEC 19795-10 standard \cite{iso19795-10}, which recommends applying the same global threshold $\tau$ to all demographic groups and computing demographic-specific False Match Rate (FMR) and False Non-Match Rate (FNMR).
We measure False Positive Differential performance (FPD), \ie, the difference in FMR, and False Negative Differential performance (FND), \ie, the difference in FNMR.

Particularly, we follow the ISO/IEC 19795-10 \cite{iso19795-10} and use the Gini coefficient to measure inequality by averaging pairwise differences between group error rates.
Given  a demographic attribute with $D$ groups $\{d_1,d_2,\ldots, d_D\}$, the Gini coefficient is defined as:
\begin{equation}
    \label{eq:spread_gini}\hspace*{-.6em}
    \mathrm{FPD} = \left(\frac{D}{D-1}\right)\frac{\sum\limits_{d_i} \sum\limits_{d_j} \bigl|\text{FMR}_{d_{i}}(\tau) - \text{FMR}_{d_{j}}(\tau)\bigr|}{2D^2 \cdot \frac{1}{D}\sum\limits_d \text{FMR}_d(\tau)}\,.\hspace*{-.2em}
\end{equation}
To compute FND, FMR in Eq.~\eqref{eq:spread_gini} is replaced by FNMR.

To be comparable with other work \cite{kotwal2024demographic,yang2021ramface,ohki2024labellessface}, we also adopt earlier metrics.
Specifically, we report the average and standard deviation of classification accuracy across demographic groups.
In this setting, a separate threshold $\tau_i$ is computed for each group $d_i$ to maximize its own accuracy, which yields overly optimistic results in terms of fairness.
While this approach does not reflect realistic deployment constraints, it remains a standard practice in the literature.
We include these metrics for completeness and direct comparison, but strongly discourage the use of such metrics in future work and emphasize that our primary analysis is in line with international fairness standards \cite{iso19795-10}.

\subsection{Datasets}
\label{sec:dataset}

Our evaluation focuses on fairness across ethnic groups, leaving other demographic attributes for future work.
We train all models on the BUPT-BalancedFace dataset \cite{wang2019racial}, which contains over 1.3 million images of more than 28k identities.
Importantly, the dataset is constructed to be balanced across $D=4$ ethnicities (\texttt{Caucasian}, \texttt{African}, \texttt{Asian}, \texttt{Indian}).
Additionally, the BUPT-GlobalFace dataset \cite{wang2019racial} is used to train models on demographically imbalanced data.

For demographic fairness analysis, we use two balanced test sets and one imbalanced test set.
The Racial Faces in the Wild (RFW) dataset \cite{wang2019racial} provides four balanced subsets corresponding to ethnic groups, and is widely adopted in fairness benchmarking as it reduces confounding factors by ensuring equal numbers of subjects and images across groups.
A few subjects in RFW are misclassified as multiple ethnicities, which we corrected.
The Balanced Faces in the Wild (BFW) dataset \cite{robinson2020face} similarly provides balanced subsets across both gender and ethnicity, serving as a complementary balanced benchmark.
RFW selects the most difficult within-ethnicity non-mated pairs, whereas BFW uses random non-mated pairs by default; for BFW, we restrict evaluation to within-ethnicity and within-gender pairs only.
For imbalanced evaluation, we use the VGGFace2 test set \cite{cao2018vggface2}, on which we perform all-vs-all comparisons across all subjects.
VGGFace2 is not balanced \wrt ethnicity and gender, for which the labels were obtained from a publicly available source,\footnote{\url{https://gitlab.idiap.ch/bob/bob.bio.face/-/blob/master/src/bob/bio/face/database/vgg2.py}} and the resulting verification pairs are likewise imbalanced.
We retain only within-group pairs, discarding cross-ethnicity and cross-gender comparisons.

To assess recognition performance at scale, we additionally evaluate on the publicly available benchmarks (LFW, CPLFW, CALFW, AgeDB, CFP-FP) and the IJB-C dataset \cite{maze2018ijbc}.
We use the same evaluation protocols for all the benchmark datasets as publicly available.\footnote{\url{https://github.com/deepinsight/insightface}}
Unlike RFW, IJB-C is not balanced demographically, but serves as a benchmark for large-scale unconstrained face verification.

\subsection{Training Setup and Backbone}
\label{sec:backbone}

We compare the proposed DeSCon variants against a broad set of state-of-the-art (SOTA) bias mitigation strategies on face recognition.
The compared methods include the baseline (ArcFace), preprocessing: DeFT \cite{kotwal2024demographic}, in-processing: RamFace \cite{yang2021ramface}, MixFairFace \cite{wang2023mixfairface}, Labelless \cite{ohki2024labellessface}, post-processing techniques: FairScoreNormalization (FSN) \cite{terhorst2020post}, Score Normalization (ScoreNorm) \cite{linghu2024score}, and our proposed DeSCon variants.
For a controlled comparison, all methods are re-implemented by us: we use official releases when available and otherwise reproduce them based on the descriptions in the original papers.
For pre- and post-processing approaches, we follow the default configurations provided in their respective papers or official source code.

Our experiments are based on models with two backbones: IResNet50 and IResNet100 \cite{deng2022arcface}.
All models are trained from scratch under an identical setup to ensure consistent comparison across backbones and bias-mitigation strategies.
Specifically, we use a scale $s=64$, margin $m=0.5$, embedding dimension of $M=512$, and batch size of $B=256$.
All models are trained for 30 epochs with an initial learning rate of 0.1, reduced by a factor of 0.1 at epochs 12 and 20.
Compared with ArcFace, DeSCon introduces additional pairwise similarity computation during training. 
In our implementation, the resulting wall-clock training time remained comparable to the ArcFace baseline (difference below 1\%), while inference incurs no additional cost since only the backbone embedding is used.

\subsection{Balanced Training Results}
\label{sec:result}
\begin{table}[t]%[!h]
  \centering
  \Caption[tab:compact]{Comprehensive Table on RFW}{
    We report the verification performance and fairness metrics for algorithms grouped by baseline, pre/post-processing, SOTA in-processing, and our proposed DeSCon methods, across IResNet50 and IResNet100 backbones.
    The metrics include TMR, Gini coefficients of FMR (FPD) and FNMR (FND), Average Accuracy (ACC), and Standard Deviation of Accuracy (STD).
    Methods with $^+$ are re-implemented by us since no source code was publicly available.
    The \textbf{best}, \underline{second-best}, and \textit{third-best} results are highlighted.
  }\vspace*{1ex}

  \footnotesize

  \subfloat[IResNet50]{
    \centering
    \resizebox{0.9\columnwidth}{!}{%
    \begin{tabular}{l|c|cc|cc}
\toprule
Method &TMR$\uparrow$ &FPD$\downarrow$ &FND$\downarrow$ &ACC$\uparrow$ &STD$\downarrow$ \\\midrule
ArcFace &74.09\% &\textit{0.6372} &0.1348 &\textit{95.43} &\textit{0.93} \\\hline
FSN &74.20\% &\textit{0.6372} &0.1378 &95.41 &\textit{0.93} \\
ScoreNorm &75.26\% &\textbf{0.5763} &\underline{0.1078} &\textit{95.43} &\textit{0.93} \\
DeFT &72.23\% &0.7582 &\textit{0.1127} &95.18 &1.04 \\\hline
RamFace$^+$ &73.08\% &0.8187 &0.1154 &\textit{95.43} &1.05 \\
MixFairFace &\textbf{77.20\%} &0.7582 &\textbf{0.0855} &\textbf{95.53} &\textbf{0.74} \\
Labelless$^+$ &73.21\% &0.7580 &0.1334 &95.08 &1.01 \\\hline
\textbf{DeSCon-All} &\textit{76.13\%} &0.6974 &0.1152 &95.33 &\underline{0.89} \\
\textbf{DeSCon-WG} &\underline{76.87\%} &0.8186 &0.1162 &\textit{95.43} &\textit{0.93} \\
\textbf{DeSCon-Hard} &74.69\% &\underline{0.5766} &0.1191 &\underline{95.49} &0.99 \\

 \bottomrule
 %    \bottomrule
    \end{tabular}%
    }
  }\vspace*{1ex}

    \subfloat[IResNet100]{
    \centering
    \resizebox{0.9\columnwidth}{!}{%
    \begin{tabular}{l|c|cc|cc}
\toprule
Method &TMR$\uparrow$ &FPD$\downarrow$ &FND$\downarrow$ &ACC$\uparrow$ &STD$\downarrow$ \\\midrule
ArcFace &75.35\% &0.8790 &0.1568 &95.99 &\textbf{0.67} \\\hline
FSN &75.65\% &0.8790 &0.1558 &95.97 &\underline{0.71} \\
ScoreNorm &76.13\% &\textit{0.6974} &0.1387 &95.99 &\textbf{0.67} \\
DeFT &76.55\% &0.9395 &0.1501 &95.89 &\textit{0.74} \\\hline
RamFace$^+$ &\textbf{78.02\%} &0.9396 &0.1602 &\underline{96.14} &\textit{0.74} \\
MixFairFace &75.56\% &\textbf{0.5766} &\textbf{0.1161} &96.02 &\textit{0.74} \\
Labelless$^+$ &\underline{77.77\%} &\underline{0.6372} &0.1603 &95.78 &0.78 \\\hline
\textbf{DeSCon-All} &\textit{77.43\%} &0.7581 &\underline{0.1262} &\textit{96.04} &0.75 \\
\textbf{DeSCon-WG} &76.98\% &0.8187 &0.1356 &\textbf{96.16} &\textbf{0.67} \\
\textbf{DeSCon-Hard} &75.78\% &\underline{0.6372} &\textit{0.1325} &96.01 &0.92 \\
 \bottomrule
 %    \bottomrule
    \end{tabular}%
    }
  }

\end{table}

\tabref{tab:compact} reports comprehensive performance and fairness metrics for all compared methods across two backbones.
We evaluate TMR as the primary performance metric, while fairness is assessed using FPD and FND.
All demographic-specific FMR/FNMR are omitted as FPD and FND already summarize the inter-group disparity in a single value.
Higher TMR indicates better verification performance, while lower values of the fairness metrics correspond to more consistent behavior across demographic groups.
% For completeness, average accuracy (ACC) and its standard deviation (STD) are also reported, though they are provided mainly for reference and will be omitted for the remaining tables; minor gaps with originally reported numbers may arise due to re-implementation differences and the cleaned RFW protocol.
For completeness, \tabref{tab:compact} also reports average accuracy (ACC) and its standard deviation (STD). 
These metrics are included mainly for reference and omitted from the remaining tables. 
Minor gaps with originally reported numbers may arise due to re-implementation differences and the cleaned RFW protocol.

From \tabref{tab:compact}, we observe an increase in ACC and a drop in STD for deeper backbones, which is consistent with the findings from the other papers \cite{kotwal2024demographic,yang2021ramface}. 
However, this is not the case for FPD or FND, where the ArcFace baseline obtains higher values when the backbone is larger.
The conclusions drawn from FPD and FND differ substantially from those based on ACC and STD, demonstrating that the latter are unreliable indicators of fairness.

Across backbones, the DeSCon methods robustly achieve higher TMR than ArcFace, with gains of about 1--3\%.
As one of the best-performing methods, MixFairFace is quite strong in mitigating bias on the FNMR side, \ie, lower FND.
The other two in-processing methods do not show such stable improvement, which may be attributed to implementation details not fully specified in the original papers. 
A similar trend can be observed for FSN and DeFT, while ScoreNorm has relatively positive performance.

Among the DeSCon variants, DeSCon-Hard reliably reduces both FPD and FND across backbones, achieving the best balance between verification performance and fairness, while DeSCon-WG and DeSCon-All reach higher TMR but show backbone-dependent FPD gains. 
All three variants consistently improve FND, as SupCon pulls mated 
pairs together regardless of sampling strategy, reducing FNMR uniformly across groups. 
However, FMR is governed by the tail of the non-mated distribution, which requires targeted regularization. 
DeSCon-Hard explicitly focuses on the hardest within-group non-mated pairs, directly shaping the tail per demographic group and leading to uniform FPD reduction, whereas DeSCon-WG and DeSCon-All sample more broadly, improving overall separation 
but lacking sufficient control over the per-group tail behavior, resulting in inconsistent FPD gains.

\begin{table}[t]%[!h]
  \centering
  \Caption[tab:DeSCon_compact2]{Comprehensive Table on BFW \& VGGFace2}{
    This table follows the same metrics as \tabref{tab:compact}.
    As fewer methods are compared, only the \textbf{best} and \underline{second-best} results are highlighted.
  }\vspace*{1ex}

  \subfloat[IResNet50]{
    \centering
    \resizebox{\columnwidth}{!}{%
    \begin{tabular}{l|c|cc|c|cc}
\toprule
&\multicolumn{3}{c|}{BFW} &\multicolumn{3}{c}{VGGFace2} \\\midrule
Method &TMR$\uparrow$ &FPD$\downarrow$ &FND$\downarrow$ &TMR$\uparrow$ &FPD$\downarrow$ &FND$\downarrow$ \\\midrule
ArcFace &76.57\% &\underline{0.4400} &0.1983 &87.03\% &0.7890 &0.1326 \\
RamFace$^+$ &\underline{77.93\%} &0.6619 &0.1899 &\textbf{87.51\%} &0.8400 &0.1335 \\
MixFairFace &\textbf{78.45\%} &0.5453 &\textbf{0.1557} &\underline{87.29\%} &0.8306 &0.1490 \\
\textbf{DeSCon-All} &77.25\% &0.5456 &\underline{0.1774} &87.14\% &0.8073 &\textbf{0.1193} \\
\textbf{DeSCon-WG} &77.55\% &0.5454 &0.1791 &87.23\% &\underline{0.7801} &0.1300 \\
\textbf{DeSCon-Hard} &77.39\% &\textbf{0.4184} &0.1777 &87.24\% &\textbf{0.7666} &\underline{0.1234} \\

 \bottomrule
 %    \bottomrule
    \end{tabular}%
    }
  }\vspace*{1ex}

    \subfloat[IResNet100]{
    \centering
    \resizebox{\columnwidth}{!}{%
    \begin{tabular}{l|c|cc|c|cc}
\toprule
&\multicolumn{3}{c|}{BFW} &\multicolumn{3}{c}{VGGFace2} \\\midrule
Method &TMR$\uparrow$ &FPD$\downarrow$ &FND$\downarrow$ &TMR$\uparrow$ &FPD$\downarrow$ &FND$\downarrow$ \\\midrule
ArcFace &77.95\% &0.6408 &0.2157 &87.85\% &0.8652 &0.1444 \\
RamFace$^+$ &78.47\% &0.5774 &0.2083 &\underline{88.24\%} &0.8045 &0.1433 \\
MixFairFace &78.35\% &0.5031 &\textbf{0.1876} &87.82\% &0.8856 &\textbf{0.1317} \\
\bf DeSCon-All &78.96\% &\underline{0.3765} &\underline{0.1907} &87.94\% &0.8010 &0.1330 \\
\bf DeSCon-WG &\underline{79.19\%} &0.5458 &0.1986 &\textbf{88.39\%} &\textbf{0.7926} &\underline{0.1325} \\
\bf DeSCon-Hard &\textbf{79.89\%} &\textbf{0.3441} &0.1935 &87.96\% &\underline{0.7975} &0.1384 \\

 \bottomrule
 %    \bottomrule
    \end{tabular}%
    }
  }

\end{table}

\tabref{tab:DeSCon_compact2} evaluates all methods on BFW and VGGFace2, which provide complementary balanced and imbalanced test conditions respectively.
Compared to RFW, BFW comes from a different distribution, reflected in the generally higher TMR values across all methods.
Despite this difference in difficulty and pair distribution, the relative performance trends largely mirror those observed on RFW, suggesting that the fairness improvements are robust across test protocols.
On IResNet50, DeSCon-Hard again stands out as the only variant that consistently reduces FPD alongside FND, while achieving TMR comparable to the baseline rather than the highest among the variants.
DeSCon-WG and DeSCon-All primarily improve FND, with FPD gains becoming more evident on IResNet100.
Overall, DeSCon demonstrates stable fairness improvements across balanced and imbalanced evaluation conditions, without sacrificing verification performance.

\subsection{Ablation Study}
\begin{table}[t]%[!h]
  \centering
  \Caption[tab:DeSCon_triplet]{Sampling Strategies and Base Loss}{
    We compare non-mated pair sampling strategies on RFW (IResNet50).
    Upper: methods under ArcFace base loss, where ArcFace + Triplet and ArcFace + MS replace the contrastive objective entirely, while DeSCon-SH and DeSCon-MS adopt alternative sampling strategies within our framework.
    Lower: DeSCon variants under AdaFace base loss, demonstrating that DeSCon is compatible with stronger margin-based losses.
    The \textbf{best} and \underline{second-best} results are highlighted separately within each base loss group.
  }\vspace*{1ex}

    \centering
    \resizebox{0.65\columnwidth}{!}{%
    \begin{tabular}{l|c|cc}
\toprule
Method &TMR$\uparrow$ &FPD$\downarrow$ &FND$\downarrow$\\
\midrule
\multicolumn{4}{l}{\textit{ArcFace Group}} \\
ArcFace &\underline{74.09\%} &0.6372 &0.1348 \\
ArcFace + MS &59.52\% &0.8186 &0.1175 \\
ArcFace + Triplet &72.41\% &0.8186 &\underline{0.1033} \\
\bf DeSCon-Hard &\textbf{74.69\%} &\underline{0.5766} &0.1191 \\
DeSCon-MS &72.33\% &0.8187 &0.1158 \\
DeSCon-SH &70.67\% &\textbf{0.5764} &\textbf{0.0888} \\

\midrule
\multicolumn{4}{l}{\textit{AdaFace Group}} \\
AdaFace &78.31\% &0.9396 &0.1209 \\
\bf DeSCon-All &78.54\% &\textbf{0.5766} &0.0873 \\
\bf DeSCon-WG &\textbf{80.14\%} &\textbf{0.5766} &\textbf{0.0835} \\
\bf DeSCon-Hard &\underline{78.78\%} &\underline{0.6372} &\underline{0.0848} \\

\bottomrule
    \end{tabular}%
    }

\end{table}

We ablate two key design dimensions of DeSCon: the choice of sampling strategy and the compatibility with alternative margin-based loss functions.
For non-mated sampling, we compare DeSCon-Hard against two alternative strategies: DeSCon-MS, which uses multi-similarity sampling \cite{wang2019multi}, and DeSCon-SH, which uses semi-hard non-mated pair sampling \cite{schroff2015facenet}.
To further contextualize the role of SupCon \eqref{eq:contrastive}, we also include two baselines that replace it entirely with other pairwise losses: ArcFace + Triplet \cite{schroff2015facenet} and ArcFace + MS \cite{wang2019multi}.
Separately, to test whether DeSCon generalizes beyond ArcFace, we replace the classification loss in Eq.~\eqref{eq:joint} with AdaFace \cite{kim2022adaface}, a stronger margin-based alternative. 
All experiments in this section are conducted on the IResNet50 backbone and evaluated on RFW.

The results are reported in \tabref{tab:DeSCon_triplet}.
Among the sampling strategies, replacing SupCon entirely (ArcFace + MS, ArcFace + Triplet) degrades TMR substantially, confirming that the SupCon cannot be simply substituted by other pairwise losses.
This can be attributed to the SoftMax-based aggregation over all non-mated pairs, which produces stronger gradients for nearby negatives than the single hardest-triplet formulation, making it more effective at shaping the tail of the non-mated score distribution.
Among the DeSCon variants, DeSCon-Hard achieves the best overall balance. %, improving both TMR and FPD over the ArcFace baseline.
DeSCon-SH achieves the lowest FND but at the cost of TMR, while DeSCon-MS offers no consistent advantage over DeSCon-Hard.
These results confirm that hard within-demographic group mining of non-mated pairs is the most effective sampling strategy within our framework.

For the AdaFace backbone, all three DeSCon variants improve fairness over the AdaFace baseline, with DeSCon-WG achieving the highest TMR and DeSCon-All and DeSCon-WG jointly achieving the lowest FPD.
Thus, DeSCon is compatible with stronger margin-based losses, and its fairness benefits are not specific to ArcFace.

\subsection{Imbalanced Training Results}
\label{sec:imbal_result}
\begin{table}[t]%[!h]
  \centering
  \Caption[tab:DeSCon_imbal]{Imbalanced Training Set}{
    The evaluation of IResNet50 models trained on the imbalanced BUPT-GlobalFace dataset follows the same metrics as \tabref{tab:compact}, on both RFW and BFW.
  }\vspace*{1ex}

    \centering
    \resizebox{\columnwidth}{!}{%
    \begin{tabular}{l|c|cc|c|cc}
\toprule
&\multicolumn{3}{c|}{RFW} &\multicolumn{3}{c}{BFW} \\\midrule
Method &TMR$\uparrow$ &FPD$\downarrow$ &FND$\downarrow$ &TMR$\uparrow$ &FPD$\downarrow$ &FND$\downarrow$ \\\midrule
ArcFace &78.52\% &0.6369 &0.0637 &74.93\% &0.8202 &0.1346 \\
RamFace$^+$ &71.83\% &0.5764 &0.0659 &\textbf{75.96\%} &0.6723 &0.1273 \\
MixFairFace &78.34\% &0.6370 &\textbf{0.0411} &74.36\% &\textbf{0.6404} &\textbf{0.1167} \\
\bf DeSCon-All &\underline{78.55\%} &0.6973 &0.0740 &75.03\% &0.7570 &0.1306 \\
\bf DeSCon-WG &77.20\% &\underline{0.5763} &0.0642 &\underline{75.37\%} &0.7885 &\underline{0.1236} \\
\bf DeSCon-Hard &\textbf{78.87\%} &\textbf{0.4553} &\underline{0.0625} &74.91\% &\underline{0.6618} &0.1279 \\

 \bottomrule
 %    \bottomrule
    \end{tabular}%
    }

\end{table}

\tabref{tab:DeSCon_imbal} reports results for IResNet50 models trained on the \emph{imbalanced} BUPT-GlobalFace dataset, evaluated on both RFW and BFW.
Compared to balanced training (\tabref{tab:compact}), training on the larger imbalanced dataset generally yields higher TMR on RFW, while TMR on BFW drops, suggesting that the effect of demographic skew in training data varies across test distributions.
Despite this, DeSCon-Hard remains the most robust variant overall, achieving the highest TMR on RFW while delivering the strongest reductions in both FPD and FND on RFW, and the second-lowest FPD on BFW.
DeSCon-WG and DeSCon-All show mixed results, with gains in some metrics offset by weaker performance in others, and neither achieves a consistent advantage over the baseline across both test sets.
MixFairFace and RamFace can mitigate bias in terms of FPD and/or FND, but at a cost of lower TMR.
Overall, DeSCon-Hard emerges as the preferred variant under training imbalance, demonstrating robust fairness gains across both test protocols.

\subsection{Benchmark Evaluation}

We additionally report benchmark evaluation results in \tabref{tab:benchmark}, focusing on the methods that achieved strong TMR in \tabref{tab:compact}.
Despite different hyperparameter settings, our ArcFace baseline maintains comparable performance across five basic benchmark datasets (LFW, CALFW, CPLFW, CFP-FP, and AgeDB) and on the IJB-C dataset with a threshold of FMR=$10^{-5}$.
The relatively low performance of MixFairFace is consistent with \cite{wang2023mixfairface}.
All three DeSCon variants maintain accuracy comparable to ArcFace across basic benchmarks, with improvements on the IJB-C dataset.
In particular, no single DeSCon variant uniformly outperforms the others across all benchmarks, while all DeSCon variants show greater relative improvements on the more challenging IJB-C dataset. %the demographic-based DeSCon variants gain more benefits when facing the challenging IJB-C.
\begin{table}[!t]%[!h]
\centering
\Caption[tab:benchmark]{Verification Benchmarks}{
  The generalization performance on the publicly available benchmarks is shown.
  We report accuracy across five benchmark datasets (LFW, CPLFW, CALFW, AgeDB, CFP-FP) and TMR at FMR of $10^{-5}$ for IJB-C, for baseline, SOTA, and proposed methods using IResNet50 and IResNet100 backbones.
  The \textbf{best}, \underline{second-best}, and \textit{third-best} results are highlighted.
}
\resizebox{\columnwidth}{!}{%
\begin{tabular}{l|ccccc|c}
\toprule
Method & LFW &  CALFW &  CPLFW & CFP\_FP & AgeDB &   IJB-C \\
\midrule
\multicolumn{7}{l}{\textit{IResNet50}} \\
ArcFace &\underline{99.67\%} &\underline{95.60\%} &91.65\% &\textbf{97.37\%} &96.78\% &90.88\% \\
DeFT &\textbf{99.72\%} &95.50\% &\underline{91.85\%} &\textit{97.24\%} &\underline{97.00\%} &90.53\% \\
RamFace\textsuperscript{+} &\textit{99.63\%} &95.53\% &\textbf{91.98\%} &\textbf{97.37\%} &\textbf{97.12\%} &\textit{91.14\%} \\
MixFairFace &\underline{99.67\%} &\textbf{95.68\%} &91.05\% &96.89\% &96.82\% &73.44\% \\
\textbf{DeSCon-All} &99.58\% &95.53\% &91.80\% &97.17\% &96.92\% &90.93\% \\
\textbf{DeSCon-WG} &\underline{99.67\%} &\textit{95.55\%} &91.72\% &\underline{97.27\%} &\textit{96.95\%} &\textbf{91.54\%} \\
\textbf{DeSCon-Hard} &99.58\% &95.40\% &\textit{91.83\%} &97.00\% &96.82\% &\underline{91.43\%} \\

\midrule
\multicolumn{7}{l}{\textit{IResNet100}} \\
ArcFace &\underline{99.68\%} &95.53\% &\textit{92.30\%} &\textit{97.79\%} &96.92\% &91.68\% \\
DeFT &99.65\% &95.45\% &92.13\% &97.71\% &\textit{97.20\%} &91.32\% \\
RamFace\textsuperscript{+} &\textbf{99.72\%} &95.42\% &92.18\% &97.63\% &\underline{97.23\%} &90.65\% \\
MixFairFace &99.63\% &\textit{95.60\%} &92.12\% &97.50\% &96.85\% &80.63\% \\
\textbf{DeSCon-All} &\textbf{99.72\%} &\textbf{95.73\%} &\underline{92.53\%} &97.71\% &97.17\% &\underline{91.95\%} \\
\textbf{DeSCon-WG} &\textit{99.67\%} &\underline{95.67\%} &\textbf{92.57\%} &\textbf{97.94\%} &\textbf{97.35\%} &\textbf{92.16\%} \\
\textbf{DeSCon-Hard} &\textbf{99.72\%} &\textit{95.60\%} &92.03\% &\underline{97.83\%} &96.85\% &\textit{91.90\%} \\
\bottomrule
\end{tabular}
}

\end{table}

\section{Conclusion}
\label{sec:conclusion}

We propose DeSCon, a fairness-aware face recognition framework that augments centroid-based margin losses with supervised contrastive learning.
% By exploring three non-mated pair sampling strategies, we show that contrastive regularization can simultaneously improve verification performance and reduce demographic disparities.
By exploring three non-mated pair sampling strategies, we show that contrastive regularization can improve the fairness-performance trade-off by reducing demographic disparities while maintaining strong verification performance.
Across multiple benchmarks, DeSCon variants achieve competitive verification performance while consistently narrowing fairness gaps, with DeSCon-Hard emerging as the most robust variant, delivering consistent improvements in both FPD and FND under balanced and imbalanced training, across test protocols covering both ethnicity-only (RFW) and joint ethnicity-gender groupings (BFW, VGGFace2).
DeSCon-WG and DeSCon-All show more mixed results, with fairness gains that are less stable across metrics and evaluation conditions.
Ablation experiments further confirm that hard negative mining is the most effective sampling strategy within our framework, and that DeSCon generalizes beyond ArcFace to stronger margin-based losses such as AdaFace while improving fairness without sacrificing verification performance.
These findings suggest that both sampling strategy and base loss selection are critical design choices for fairness-aware training.

Importantly, our evaluation highlights the necessity of adhering to \emph{standardized fairness metrics}, such as the Gini coefficient, to provide a rigorous assessment of bias mitigation methods.
Through re-implementing related work, we provide a more application-oriented fairness evaluation of existing methods, which serves as a basis for future comparison.

% To ensure comparability with related work and to keep training costs low, we trained all our methods solely on the ethnicity-labeled but only moderate-scale BUPT datasets.
% This comes at the cost of lower overall verification performance as compared to training with larger-scale datasets such as WebFace-260M \cite{zhu2021webface260m}, which lacks reliable demographic annotations.
% Extending DeSCon to WebFace-scale training with estimated demographics remains an important direction for future work.

To ensure comparability with related work and to keep training costs low, we trained all methods on the moderate-scale BUPT datasets, which provide the ethnicity annotations required by DeSCon during training.
This choice comes at the cost of lower overall verification performance compared to training on larger-scale datasets such as WebFace-260M \cite{zhu2021webface260m}, which currently lacks reliable demographic annotations, although DeSCon does not require demographic information at inference time.
Extending DeSCon to WebFace-scale training with estimated or automatically inferred demographics remains an important direction for future work.

\section*{Acknowledgement}
The authors thank the Hasler foundation for
their support through the SAFER project.

{\small
\bibliographystyle{ieee}
\bibliography{texfiles/References,texfiles/Publications}

@article{mandasari2014calibration,
  title={Score calibration in face recognition},
  author={Mandasari, Miranti Indar and G\"unther, Manuel and Wallace, Roy and Saeidi, Rahim and Marcel, S{\'e}bastien and van Leeuwen, David A.},
  journal={IET Biometrics},
  volume={3},
  number={4},
  optpages={246--256},
  year={2014},
  publisher={IET}
}

@inbook{rudd2018hiding,
  author={Rudd, Ethan M. and G\"unther, Manuel and Dhamija, Akshay Raj and Kateb, Faris A. and Boult, Terrance E.},
  title={What's Hiding in My Deep Features?},
  year=2018,
  booktitle={Deep Learning in Biometrics},
  publisher={CRC Press}
}

@article{pereira20228years,
  title   = {Eight Years of Face Recognition Research: Reproducibility, Achievements and Open Issues},
  author  = {de Freitas Pereira, Tiago and Schmidli, Dominic and Linghu, Yu and Zhang, Xinyi and Marcel, S\'ebastien and G\"unther, Manuel},
  year    = {2022},
  journal = {arXiv}
}

@inproceedings{palechor2023protocols,
  author    = {Palechor, Andres and Bhoumik, Annesha and G\"unther, Manuel},
  title     = {Large-Scale Open-Set Classification Protocols for {ImageNet}},
  booktitle = {Winter Conference on Applications of Computer Vision (WACV)},
  year      = {2023},

}

@inproceedings{kasim2024challenge,
  title     = {Watchlist Challenge: 3rd Open-set Face Detection and Identification},
  author    = {Kasim, Furkan and Boult, Terrance E. and  Mora, Rensso and Biesseck, Bernardo and Ribeiro, Rafael and Schlueter, Jan and Repak, Tomáš and Vareto, Rafael Henrique and  Menotti, David  and Schwartz, William Robson and  G\"unther, Manuel},
  booktitle = {International Joint Conference on Biometrics (IJCB)},
  year      = {2024},
}

@inproceedings{linghu2024score,
  title     = {Score Normalization for Demographic Fairness in Face Recognition},
  author    = {Linghu, Yu and de Freitas Pereira, Tiago and Ecabert, Christophe and Marcel, S\'ebastien and G\"unther, Manuel},
  booktitle = {International Joint Conference on Biometrics (IJCB)},
  year      = {2024},
}

@inproceedings{schroff2015facenet,
  title={Facenet: A unified embedding for face recognition and clustering},
  author={Schroff, Florian and Kalenichenko, Dmitry and Philbin, James},
  booktitle={Conference on Computer Vision and Pattern Recognition (CVPR)},
  optpages={815--823},
  year={2015}
}

@inproceedings{cao2018vggface2,
author    = {Cao, Qiong and Shen, Li and Xie, Weidi and Parkhi, Omkar M. and Zisserman, Andrew},
title     = {{VGGFace2}: A Dataset for Recognising Faces across Pose and Age},
booktitle = {Automatic Face {\&} Gesture Recognition (FG)},
year      = {2018},
optorganization={IEEE}
}

@article{otoole2007surpass,
  author = {O'Toole, Alice J. and Phillips, P. Jonathon and Jiang, Fang and Ayyad, Janet H. and Penard, Nils and Abdi, Herv\`e},
  journal = {Transactions on Pattern Analysis and Machine Intelligence (TPAMI)},
  optnumber = 9,
  optpages = {1642-1646},
  title = {Face Recognition Algorithms Surpass Humans Matching Faces Over Changes in Illumination},
  optvolume = 29,
  year = 2007
}

@inproceedings{maze2018ijbc,
  author={Maze, Brianna and Adams, Jocelyn and Duncan, James A. and Kalka, Nathan and Miller, Tim and Otto, Charles and Jain, Anil K.  and Niggel, W. Tyler and Anderson, Janet and Cheney, Jordan and  Grother, Patrick},
  booktitle={International Conference on Biometrics (ICB)},
  title={{IARPA} {Janus} {Benchmark - C}: Face Dataset and Protocol},
  year={2018},
}

@inproceedings{he2016deep,
  title={Deep Residual Learning for Image Recognition},
  author={He, Kaiming and Zhang, Xiangyu and Ren, Shaoqing and Sun, Jian},
  booktitle={Conference on Computer Vision and Pattern Recognition (CVPR)},
  year={2016},
  optorganization={IEEE}
}

@inproceedings{hu2018squeeze,
  title={Squeeze-and-Excitation Networks},
  author={Hu, Jie and Shen, Li and Sun, Gang},
  booktitle={Conference on Computer Vision and Pattern Recognition (CVPR)},
  year={2018}
}

@article{hill2020wrongfully,
  author = {Hill, Kashmir},
  title = {Wrongfully Accused by an Algorithm},
  journal = {New York Times},
  year = {2020},
  optmonth = {6},
  optday = {24},
  note = {\href{https://www.nytimes.com/2020/06/24/technology/facial-recognition-arrest.html}{https://www.nytimes.com}}
}

@article{wallace2012cross,
  author   = {Wallace, Roy and McLaren, Mitchell and McCool, Chris and Marcel, S\'ebastien},
  title    = {Cross-pollination of normalisation techniques from speaker to face authentication using Gaussian mixture models},
  journal  = {Transactions on Information Forensics and Security (TIFS)},
  year     = {2012}
}

@inproceedings{meng2021magface,
  title     = {{MagFace}: A universal representation for face recognition and quality assessment},
  author    = {Meng, Qiang and Zhao, Shichao and Huang, Zhida and Zhou, Feng},
  booktitle = {Conference on Computer Vision and Pattern Recognition (CVPR)},
  year      = {2021}
}

@inproceedings{zhu2021webface260m,
  title     = {{WebFace260M}: A Benchmark Unveiling the Power of Million-scale Deep Face Recognition},
  author    = {Zhu, Zheng and Huang, Guan and Deng, Jiankang and Ye, Yun and Huang, Junjie and Chen, Xinze and Zhu, Jiagang and Yang, Tian and Lu, Jiwen and Du, Dalong and others},
  booktitle = {Conference on Computer Vision and Pattern Recognition (CVPR)},
  year      = {2021}
}

@inproceedings{zhang2019adacos,
  title     = {{AdaCos}: Adaptively Scaling Cosine Logits for Effectively Learning Deep Face Representations},
  author    = {Zhang, Xiao and Zhao, Rui and Qiao, Yu and Wang, Xiaogang and Li, Hongsheng},
  year      = {2019},
  booktitle = {Conference on Computer Vision and Pattern Recognition (CVPR)},
  optpages  = {10823–10832}
}

@inproceedings{wang2019racial,
  title        = {Racial Faces in the Wild: Reducing Racial Bias by Information Maximization Adaptation Network},
  booktitle    = {International Conference on Computer Vision (ICCV)},
  author       = {Wang, Mei and Deng, Weihong and Hu, Jiani and Tao, Xunqiang and Huang, Yaohai},
  year         = {2019},
  optorganization = {IEEE},
  optpages        = {692--702}
}

@article{krishnapriya2020skintone,
  author  = {Krishnapriya, K. S. and Albiero, Vítor and Vangara, Kushal and King, Michael C. and Bowyer, Kevin W.},
  journal = {Transactions on Technology and Society (TTS)},
  title   = {Issues Related to Face Recognition Accuracy Varying Based on Race and Skin Tone},
  year    = {2020},
  optvolume  = {1},
  optnumber  = {1},
  optpages   = {8--20},
  doi     = {10.1109/TTS.2020.2974996}
}

@inproceedings{albiero2020gender,
  title        = {Analysis of gender inequality in face recognition accuracy},
  author       = {Albiero, Vitor and KS, Krishnapriya and Vangara, Kushal and Zhang, Kai and King, Michael C and Bowyer, Kevin W},
  booktitle    = {Winter Conference on Applications of Computer Vision Workshops (WACVW)},
  optpages        = {81--89},
  optorganization = {IEEE/CVF},
  year         = {2020}
}

@article{cavazos2020accuracy,
  title     = {Accuracy comparison across face recognition algorithms: Where are we on measuring race bias?},
  author    = {Cavazos, Jacqueline G and Phillips, P Jonathon and Castillo, Carlos D and O’Toole, Alice J},
  journal   = {Transactions on Biometrics, Behavior, and Identity Science (TBIOM)},
  optvolume    = {3},
  optnumber    = {1},
  optpages     = {101--111},
  year      = {2020},
  publisher = {IEEE}
}

@inproceedings{michalski2018impact,
  title        = {The impact of age and threshold variation on facial recognition algorithm performance using images of children},
  author       = {Michalski, Dana and Yiu, Sau Yee and Malec, Chris},
  booktitle    = {International Conference on Biometrics (ICB)},
  optpages        = {217--224},
  year         = {2018},
  optorganization = {IEEE}
}

@inproceedings{albiero2020balance,
  author    = {Albiero, Vítor and Zhang, Kai and Bowyer, Kevin W.},
  booktitle = {International Joint Conference on Biometrics (IJCB)},
  title     = {How Does Gender Balance In Training Data Affect Face Recognition Accuracy?},
  year      = {2020},
}

@article{cook2019demographic,
  title     = {Demographic effects in facial recognition and their dependence on image acquisition: An evaluation of eleven commercial systems},
  author    = {Cook, Cynthia M and Howard, John J and Sirotin, Yevgeniy B and Tipton, Jerry L and Vemury, Arun R},
  journal   = {Transactions on Biometrics, Behavior, and Identity Science (TBIOM)},
  optvolume    = {1},
  optnumber    = {1},
  optpages     = {32--41},
  year      = {2019},
  publisher = {IEEE}
}

@inproceedings{poh2010group,
  title        = {Group-specific score normalization for biometric systems},
  author       = {Poh, Norman and Kittler, Josef and Rattani, Ajita and Tistarelli, Massimo},
  booktitle    = {Conference on Computer Vision and Pattern Recognition Workshops (CVPRW)},
  optpages        = {38--45},
  year         = {2010},
  optorganization = {IEEE}
}

@article{pereira2021fairness,
  author  = {de Freitas Pereira, Tiago and Marcel, S\'ebastien},
  journal = {Transactions on Biometrics, Behavior, and Identity Science (TBIOM)},
  title   = {Fairness in Biometrics: A Figure of Merit to Assess Biometric Verification Systems},
  year    = {2021},
  optvolume  = {4},
  optnumber  = {1},
  optpages   = {19-29},
  optpublisher={IEEE}
}

@inproceedings{robinson2020face,
  title     = {Face recognition: too bias, or not too bias?},
  author    = {Robinson, Joseph P and Livitz, Gennady and Henon, Yann and Qin, Can and Fu, Yun and Timoner, Samson},
  booktitle = {Computer Vision and Pattern Recognition Workshops (CVPRW)},
  year      = {2020}
}

@article{robinson2023balancing,
  title={Balancing biases and preserving privacy on balanced faces in the wild},
  author={Robinson, Joseph P and Qin, Can and Henon, Yann and Timoner, Samson and Fu, Yun},
  journal={IEEE Transactions on Image Processing (TIP)},
  year={2023},
  publisher={IEEE}
}

@inproceedings{kim2022adaface,
  title     = {{AdaFace}: Quality Adaptive Margin for Face Recognition},
  author    = {Kim, Minchul and Jain, Anil K. and Liu, Xiaoming},
  booktitle = {Conference on Computer Vision and Pattern Recognition (CVPR)},
  year      = {2022}
}

@inproceedings{duta2021iresnet,
  title        = {Improved residual networks for image and video recognition},
  author       = {Duta, Ionut Cosmin and Liu, Li and Zhu, Fan and Shao, Ling},
  booktitle    = {International Conference on Pattern Recognition (ICPR)},
  year         = {2021},
  optorganization = {IEEE}
}

@techreport{grother2022nist,
  author      = {Grother, Patrick},
  institution = {National Institute of Standards and Technology (NIST)},
  title       = {Face Recognition Vendor Test {(FRVT)} Part 8: Summarizing Demographic Differentials},
  year        = {2022}
}

@inproceedings{vangara2019characterizing,
  title     = {Characterizing the variability in face recognition accuracy relative to race},
  booktitle = {Conference on Computer Vision and Pattern Recognition Workshops (CVPRW)},
  author    = {Vangara, Kushal and King, Michael C and Albiero, Vitor and Bowyer, Kevin and others},
  year      = {2019}
}

@article{huang2019deep,
  title={Deep imbalanced learning for face recognition and attribute prediction},
  author={Huang, Chen and Li, Yining and Loy, Chen Change and Tang, Xiaoou},
  journal={Transactions on Pattern Analysis and Machine Intelligence (TPAMI)},
  year={2019},
  publisher={IEEE}
}

@inproceedings{kortylewski2019analyzing,
  title     = {Analyzing and reducing the damage of dataset bias to face recognition with synthetic data},
  author    = {Kortylewski, Adam and Egger, Bernhard and Schneider, Andreas and Gerig, Thomas and Morel-Forster, Andreas and Vetter, Thomas},
  booktitle = {Conference on Computer Vision and Pattern Recognition (CVPR)},
  year      = {2019}
}

@inproceedings{wang2020mitigating,
  title     = {Mitigating Bias in Face Recognition Using Skewness-Aware Reinforcement Learning},
  booktitle = {Conference on Computer Vision and Pattern Recognition (CVPR)},
  author    = {Wang, Mei and Deng, Weihong},
  year      = {2020}
}

@article{wang2019deep,
  title     = {Deep class-skewed learning for face recognition},
  author    = {Wang, Pingyu and Su, Fei and Zhao, Zhicheng and Guo, Yandong and Zhao, Yanyun and Zhuang, Bojin},
  journal   = {Neurocomputing},
  year      = {2019},
  publisher = {Elsevier}
}

@inproceedings{gong2020jointly,
  title        = {Jointly de-biasing face recognition and demographic attribute estimation},
  author       = {Gong, Sixue and Liu, Xiaoming and Jain, Anil K},
  booktitle    = {European Conference on Computer Vision (ECCV)},
  year         = {2020},
  optorganization = {Springer}
}

@inproceedings{liang2019additive,
  title={Additive adversarial learning for unbiased authentication},
  author={Liang, Jian and Cao, Yuren and Zhang, Chenbin and Chang, Shiyu and Bai, Kun and Xu, Zenglin},
  booktitle={Conference on Computer Vision and Pattern Recognition (CVPR)},
  year={2019}
}

@article{morales2020sensitivenets,
  title={{SensitiveNets}: Learning agnostic representations with application to face images},
  author={Morales, Aythami and Fierrez, Julian and Vera-Rodriguez, Ruben and Tolosana, Ruben},
  journal={Transactions on Pattern Analysis and Machine Intelligence (TPAMI)},
  year={2020},
  publisher={IEEE}
}

@inproceedings{gong2021mitigating,
  title={Mitigating face recognition bias via group adaptive classifier},
  author={Gong, Sixue and Liu, Xiaoming and Jain, Anil K},
  booktitle={Conference on Computer Vision and Pattern Recognition (CVPR)},
  year={2021}
}

@article{serna2022sensitive,
  title     = {Sensitive loss: Improving accuracy and fairness of face representations with discrimination-aware deep learning},
  author    = {Serna, Ignacio and Morales, Aythami and Fierrez, Julian and Obradovich, Nick},
  journal   = {Artificial Intelligence},
  year      = {2022},
  publisher = {Elsevier}
}

@inproceedings{yang2021ramface,
  title={{RamFace}: Race adaptive margin based face recognition for racial bias mitigation},
  author={Yang, Zhanjia and Zhu, Xiangping and Jiang, Changyuan and Liu, Wenshuang and Shen, Linlin},
  booktitle={International Joint Conference on Biometrics (IJCB)},
  year={2021},
  optorganization={IEEE}
}

@inproceedings{terhoerst2020comparison,
  title        = {Comparison-level mitigation of ethnic bias in face recognition},
  author       = {Terh{\"o}rst, Philipp and Tran, Mai Ly and Damer, Naser and Kirchbuchner, Florian and Kuijper, Arjan},
  booktitle    = {International Workshop on Biometrics and Forensics (IWBF)},
  year         = {2020},
  optorganization = {IEEE}
}

@inproceedings{wu2024facial-hair,
  author    = {Wu, Haiyu and Tian, Sicong and Bhatta, Aman and \"Ozt\"urk, Ka\u{g}an and Ricanek, Karl and Bowyer, Kevin W.},
  title     = {Facial Hair Area in Face Recognition Across Demographics: Small Size, Big Effect},
  booktitle = {Winter Conference on Applications of Computer Vision Workshops (WACVW)},
  year      = {2024},
}

@inproceedings{kotwal2024mitigating,
  title={Mitigating Demographic Bias in Face Recognition via Regularized Score Calibration},
  author={Kotwal, Ketan and Marcel, S{\'e}bastien},
  booktitle={Winter Conference on Applications of Computer Vision (WACV)},
  year={2024}
}

@article{deng2022arcface,
  author   = {Deng, Jiankang and Guo, Jia and Yang, Jing and Xue, Niannan and Kotsia, Irene and Zafeiriou, Stefanos},
  journal  = {Transactions on Pattern Analysis and Machine Intelligence (TPAMI)},
  title    = {ArcFace: Additive Angular Margin Loss for Deep Face Recognition},
  year     = {2022},
  doi      = {10.1109/TPAMI.2021.3087709}
}

@inproceedings{terhorst2020beyond,
  title        = {Beyond identity: What information is stored in biometric face templates?},
  author       = {Terh{\"o}rst, Philipp and F{\"a}hrmann, Daniel and Damer, Naser and Kirchbuchner, Florian and Kuijper, Arjan},
  booktitle    = {International Joint Conference on Biometrics (IJCB)},
  year         = {2020},
}

@article{terhorst2020post,
  title     = {Post-comparison mitigation of demographic bias in face recognition using fair score normalization},
  author    = {Terh{\"o}rst, Philipp and Kolf, Jan Niklas and Damer, Naser and Kirchbuchner, Florian and Kuijper, Arjan},
  journal   = {Pattern Recognition Letters},
  year      = {2020},
}

@inproceedings{robbins2023cast,
  title     = {{CAST}: Conditional attribute subsampling toolkit for fine-grained evaluation},
  author    = {Robbins, Wes and Zhou, Steven and Bhatta, Aman and Mello, Chad and Albiero, V{\'\i}tor and Bowyer, Kevin W and Boult, Terrance E},
  booktitle = {Winter Conference on Applications of Computer Vision (WACV)},
  year      = {2023}
}

@article{hidayat2024abc,
  author   = {Hidayat, Fadhil and Elviani, Ulva and Situmorang, George Bryan Gabriel and Ramadhan, Muhammad Zaky and Alunjati, Figo Agil and Sucipto, Reza Fauzi},
  journal  = {IEEE Access},
  title    = {Face Recognition for Automatic Border Control: A Systematic Literature Review},
  year     = {2024},
  doi      = {10.1109/ACCESS.2024.3373264}
}

@article{kunda2021android,
  title     = {A survey of android mobile phone authentication schemes},
  author    = {Kunda, Douglas and Chishimba, Mumbi},
  journal   = {Mobile Networks and Applications},
  year      = {2021},
}

@inproceedings{xu2021consistent,
  title   = {Consistent Instance False Positive Improves Fairness in Face Recognition},
  author  = {Xu, Xingkun and Huang, Yuge and Shen, Pengcheng and Li, Shaoxin and Li, Jilin and Huang, Feiyue and Li, Yong and Cui, Zhen},
  year    = {2021},
  booktitle={Conference on Computer Vision and Pattern Recognition (CVPR)},
}

@inproceedings{sun2022part,
  title   = {Part-based face recognition with vision transformers},
  author  = {Sun, Zhonglin and Tzimiropoulos, Georgios},
  booktitle = {British Machine Vision Conference (BMVC)},
  year    = {2022}
}

@inproceedings{zhang2017age,
  title={Age progression/regression by conditional adversarial autoencoder},
  author={Zhang, Zhifei and Song, Yang and Qi, Hairong},
  booktitle={Conference on Computer Vision and Pattern Recognition (CVPR)},
  year={2017}
}

@inproceedings{karkkainen2021fairface,
  title={Fairface: Face attribute dataset for balanced race, gender, and age for bias measurement and mitigation},
  author={Karkkainen, Kimmo and Joo, Jungseock},
  booktitle={Winter Conference on Applications of Computer Vision (WACV)},
  year={2021}
}

@inproceedings{ardeshir2022estimating,
  title={Estimating structural disparities for face models},
  author={Ardeshir, Shervin and Segalin, Cristina and Kallus, Nathan},
  booktitle={Conference on Computer Vision and Pattern Recognition (CVPR)},
  year={2022}
}

@inproceedings{jung2022learning,
  title={Learning fair classifiers with partially annotated group labels},
  author={Jung, Sangwon and Chun, Sanghyuk and Moon, Taesup},
  booktitle={Conference on Computer Vision and Pattern Recognition (CVPR)},
  year={2022}
}

@inproceedings{kolla2023impact,
  title={The impact of racial distribution in training data on face recognition bias: A closer look},
  author={Kolla, Manideep and Savadamuthu, Aravinth},
  booktitle={Winter Conference on Applications of Computer Vision (WACV)},
  year={2023}
}

@article{nagpal2022detox,
  title={Detox Loss: Fairness Constraints for Learning With Imbalanced Data},
  author={Nagpal, Shruti and Singh, Maneet and Singh, Richa and Vatsa, Mayank},
  journal={IEEE Transactions on Biometrics, Behavior, and Identity Science (TBIOM)},
  year={2022},
}

@article{iurada2024fairness,
  title={Fairness Meets Cross-Domain Learning: A Benchmark of Models and Metrics},
  author={Iurada, Leonardo and Bucci, Silvia and Hospedales, Timothy M and Tommasi, Tatiana},
  journal={IEEE Access},
  year={2024},
}

@inproceedings{khosla2020supervised,
  title={Supervised contrastive learning},
  author={Khosla, Prannay and Teterwak, Piotr and Wang, Chen and Sarna, Aaron and Tian, Yonglong and Isola, Phillip and Maschinot, Aaron and Liu, Ce and Krishnan, Dilip},
  booktitle={Advances in Neural Information Processing Systems (NeurIPS)},
  year={2020}
}

@inproceedings{du2022fairdisco,
  title={{FairDisco}: Fairer {AI} in dermatology via disentanglement contrastive learning},
  author={Du, Siyi and Hers, Ben and Bayasi, Nourhan and Hamarneh, Ghassan and Garbi, Rafeef},
  booktitle={European Conference on Computer Vision (ECCV)},
  year={2022},
}

@misc{iso19795-10,
  author = {{International Electrotechnical Commission (IEC)}},
  title = {{ISO/IEC 19795-10 Biometric performance testing and reporting -- Part 10}: Quantifying biometric system
performance variation across demographic groups},
  year = 2024,
  note = {first edition},
  url = {https://webstore.iec.ch/en/publication/101876}
}

@inproceedings{wang2023mixfairface,
  title={Mixfairface: Towards ultimate fairness via mixfair adapter in face recognition},
  author={Wang, Fu-En and Wang, Chien-Yi and Sun, Min and Lai, Shang-Hong},
  booktitle={AAAI Conference on Artificial Intelligence},
  year={2023}
}

@inproceedings{ohki2024labellessface,
  title={Labellessface: Fair metric learning for face recognition without attribute labels},
  author={Ohki, Tetsushi and Sato, Yuya and Nishigaki, Masakatsu and Ito, Koichi},
  booktitle={International Joint Conference on Biometrics (IJCB)},
  year={2024},
}

@inproceedings{kotwal2024demographic,
  title={Demographic fairness transformer for bias mitigation in face recognition},
  author={Kotwal, Ketan and Marcel, S{\'e}bastien},
  booktitle={International Joint Conference on Biometrics (IJCB)},
  year={2024},
}

@inproceedings{chen2020contrastive,
  title        = {A simple framework for contrastive learning of visual representations},
  author       = {Chen, Ting and Kornblith, Simon and Norouzi, Mohammad and Hinton, Geoffrey},
  booktitle    = {International Conference on Machine Learning (ICML)},
  year         = {2020},
}

@article{lin2024improving,
  title={Improving fairness of automated chest radiograph diagnosis by contrastive learning},
  author={Lin, Mingquan and Li, Tianhao and Sun, Zhaoyi and Holste, Gregory and Ding, Ying and Wang, Fei and Shih, George and Peng, Yifan},
  journal={Radiology: Artificial Intelligence},
  year={2024},
  publisher={Radiological Society of North America}
}

@inproceedings{park2022fair,
  title={Fair contrastive learning for facial attribute classification},
  author={Park, Sungho and Lee, Jewook and Lee, Pilhyeon and Hwang, Sunhee and Kim, Dohyung and Byun, Hyeran},
  booktitle={Conference on Computer Vision and Pattern Recognition (CVPR)},
  year={2022}
}

@article{ahn2024uncertainty,
  title={Uncertainty-aware face embedding with contrastive learning for open-set evaluation},
  author={Ahn, Kyeongjin and Lee, Seungeon and Han, Sungwon and Low, Cheng Yaw and Cha, Meeyoung},
  journal={IEEE Transactions on Information Forensics and Security (TIFS)},
  year={2024},
}

@inproceedings{liu2023pose,
  title={Pose-disentangled contrastive learning for self-supervised facial representation},
  author={Liu, Yuanyuan and Wang, Wenbin and Zhan, Yibing and Feng, Shaoze and Liu, Kejun and Chen, Zhe},
  booktitle={Conference on Computer Vision and Pattern Recognition (CVPR)},
  year={2023}
}

@inproceedings{ramachandran2024self,
  title={A self-supervised learning pipeline for demographically fair facial attribute classification},
  author={Ramachandran, Sreeraj and Rattani, Ajita},
  booktitle={International Joint Conference on Biometrics (IJCB)},
  year={2024},
}

@article{song2024coreface,
  title={Coreface: Sample-guided contrastive regularization for deep face recognition},
  author={Song, Youzhe and Wang, Feng},
  journal={Pattern Recognition},
  year={2024},
}

@inproceedings{wang2019multi,
  title={Multi-similarity loss with general pair weighting for deep metric learning},
  author={Wang, Xun and Han, Xintong and Huang, Weilin and Dong, Dengke and Scott, Matthew R},
  booktitle={Conference on Computer Vision and Pattern Recognition (CVPR)},
  year={2019}
}
}

\end{document}